\documentclass[10pt,twocolumn,letterpaper]{article}

\usepackage{cvpr}
\usepackage{times}
\usepackage{epsfig}
\usepackage{graphicx}
\usepackage{amsmath}
\usepackage{amssymb}
\usepackage{color}

\usepackage[utf8]{inputenc} 
\usepackage[T1]{fontenc}    
\usepackage{url}            
\usepackage{booktabs}       
\usepackage{amsfonts}       
\usepackage{nicefrac}       
\usepackage{microtype}      
\usepackage{cleveref}
\usepackage[utf8]{inputenc}
\usepackage{caption}
\usepackage{subcaption}
\usepackage{bbm}
\usepackage{algorithm,algorithmic}
\usepackage[toc,page]{appendix}
\usepackage{adjustbox}

\usepackage[pagebackref=true,breaklinks=true,letterpaper=true,colorlinks,bookmarks=false]{hyperref}

\cvprfinalcopy 

\definecolor{darkgreen}{RGB}{10,180,10}
\definecolor{darkblue}{RGB}{10,10,180}

\begin{document}

\title{Multi-Content GAN for Few-Shot Font Style Transfer } 

\author{Samaneh Azadi$^{1}$\thanks{Work done during an internship at Adobe Research}, Matthew Fisher$^{2}$, Vladimir Kim$^{2}$, Zhaowen Wang$^{2}$, Eli Shechtman$^{2}$, Trevor Darrell$^{1}$\\
$^{1}$UC Berkeley, $^2$Adobe Research\\
{\hspace{-1.5cm}\tt\small \{sazadi,trevor\}@eecs.berkeley.edu} ${\qquad}$ {\tt\small\{matfishe,vokim,zhawang,elishe\}@adobe.com}}

\maketitle

\begin{abstract}

In this work, we focus on the challenge of taking partial observations of highly-stylized text and generalizing the observations to generate unobserved glyphs in the ornamented typeface. To generate a set of multi-content images following a consistent style from very few examples, we propose an end-to-end stacked conditional GAN model considering content along channels and style along network layers. Our proposed network transfers the style of given glyphs to the contents of unseen ones, capturing highly stylized fonts found in the real-world such as those on movie posters or infographics. We seek to transfer both the typographic stylization (ex. serifs and ears) as well as the textual stylization (ex. color gradients and effects.) We base our experiments on our collected data set including 10,000 fonts with different styles and demonstrate effective generalization from a very small number of observed glyphs.
\end{abstract}

\section{Introduction}
Text is a prominent visual element of 2D design.  Artists invest significant time into designing glyphs that are visually compatible with other elements in their shape and texture. This process is labor intensive and artists often design only the subset of glyphs that are necessary for a title or an annotation, which makes it difficult to alter the text after the design is created, or to transfer an observed instance of a font to your own project. In this work, we propose a neural network architecture that automatically synthesizes the missing glyphs from a few image examples.

Early research on glyph synthesis focused on geometric modeling of outlines~\cite{suveeranont2010example,campbell2014learning,phan2015flexyfont}, which is limited to particular glyph topology (e.g., cannot be applied to decorative or hand-written glyphs) and cannot be used with image input. With the rise of deep neural networks, researchers have looked at modeling glyphs from images~\cite{baluja2016learning,upchurch2016z,lyu2017auto,chang2017chinese}. We improve this approach by leveraging recent advances in conditional generative adversarial networks (cGANS)~\cite{isola2016image}, which have been successful in many generative applications, but produce significant artifacts when directly used to generate fonts (Figure~\ref{fig:ablation}, 2nd row). Instead of training a single network for all possible typeface ornamentations, we show how to use our multi-content GAN architecture to retrain a customized network for each observed character set with only a handful of observed glyphs.

Our network operates in two stages, first modeling the overall glyph shape and then synthesizing the final appearance with color and texture, enabling transfer of fine decorative elements. Some recent texture transfer techniques directly leverage glyph structure as guiding channels to improve the placement of decorative elements~\cite{yang2016awesome}. While this approach provides good results on clean glyphs it tends to fail on automatically-generated glyphs, as the artifacts of the synthesis procedure make it harder to obtain proper guidance from the glyph structure. Instead, we propose to train an ornamentation network jointly with the glyph generation network, enabling our ornament synthesis approach to learn how to decorate automatically generated glyphs with color and texture and also fix issues that arise during glyph generation. We demonstrate that users strongly preferred the output of our glyph ornamentation network in the end-to-end glyph synthesis pipeline.


\textbf{Our Contributions.} In this paper, we propose the first end-to-end solution to synthesizing ornamented glyphs from images of a few example glyphs in the same style. To enable this, we develop a novel stacked cGAN architecture to predict the coarse glyph shapes, and a novel ornamentation \textrm{network to predict color and texture of the final glyphs.} These networks are trained jointly and specialized for each typeface using a very small number of observations, and we demonstrate the benefit of each component in our architecture (Figure~\ref{fig:ablation}). We use a perceptual evaluation to demonstrate the benefit of our jointly-trained network over effect transfer approaches augmented with a baseline glyph-outline inference network (Section~\ref{sec:perceptualEvaluation}).

Our Multi-Content GAN (\textit{MC-GAN}) code and dataset are available at \url{https://github.com/azadis/MC-GAN}. 

\begin{figure}[t!]
\centering
\includegraphics[width=\linewidth]{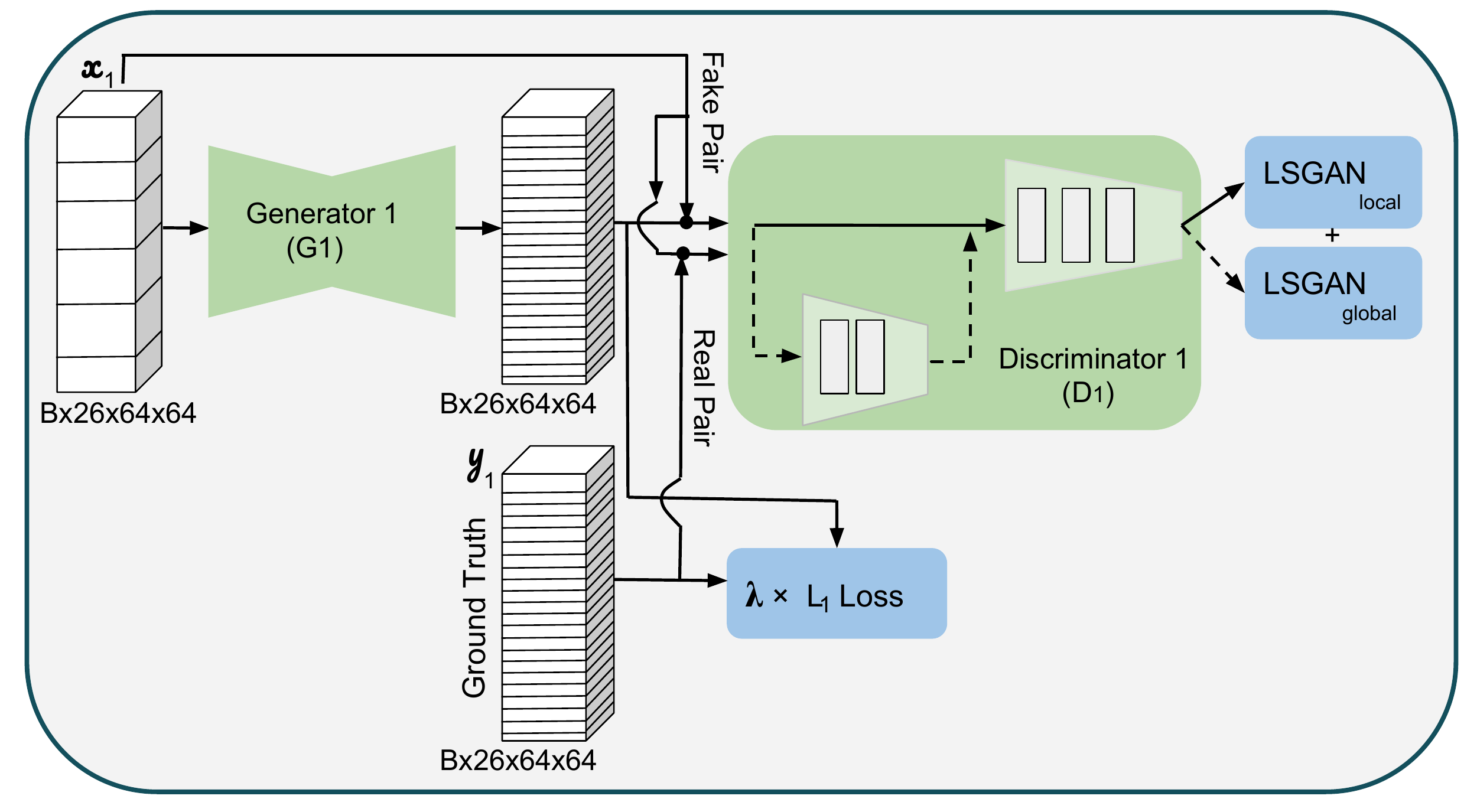}
\caption{Schematic of our Glyph Network to be trained on our $10\mathrm{K}$ font data set.}
\label{fig:glyph}
\end{figure}

\section{Related Work}

Font glyph synthesis from few examples has been a long-studied problem. Earlier methods~\cite{suveeranont2010example,campbell2014learning,phan2015flexyfont} mostly relied on explicit shape modeling to construct the transformation between existing and novel glyphs. Glyph part models for radicals~\cite{zhou2011easy} and strokes~\cite{lian2016automatic} were designed specifically for Chinese characters. Based on a shape representation, machine learning techniques, including statistical models~\cite{phan2015flexyfont} and bilinear factorization~\cite{tenenbaum1997separating}, have been used to infer and transfer stroke styles and composition rules. More recently, with the rise of deep learning, convolutional neural networks have also been applied to novel glyph synthesis. Promising results were obtained with conventional model structures~\cite{baluja2016learning,upchurch2016z} as well as generative adversarial networks (GANs)~\cite{lyu2017auto,chang2017chinese}. All these networks only predict glyph shape, a goal also targeted by our glyph network. We adopt a distinct multi-content representation in our glyph network which proves to effectively capture the common style among multiple glyphs.

Transferring artistic styles of color and texture to new glyphs is a challenging problem distinct from inferring the overall glyph shape. The problem was investigated in~\cite{yang2016awesome} with the assumption that the unstylized glyph shape is given. A patch-based texture synthesis algorithm is employed to map sub-effect patterns to correlated positions on text skeleton for effect generation. Style transfer has been more actively studied on general images with the aid of convolutional neural networks (CNNs). CNN features are successfully used to represent image styles, and serve as the basis for optimization~\cite{gatys2016image,li2016combining,liao2017visual}. Recently, networks trained with feed-forward structure and adversarial loss have achieved much improved efficiency~\cite{li2016precomputed,johnson2016perceptual} and generalization ability~\cite{huang2017arbitrary,li2017universal}. Our proposed ornamentation network is the first to employ deep networks for text effect transfer.

Several problems in graphics and vision require synthesizing data that is consistent with partial observations. These methods typically focus on learning domain-specific priors to accomplish this task. For example, given a single-view image, encoder-decoder architectures have been proposed to hallucinate novel views of faces~\cite{kulkarni2015deep,tran2017disentangled}, bodies~\cite{zhao2017multi}, and other rigid objects~\cite{zhou2016view,park2017transformation}. CNNs were also used to complete missing regions in images~\cite{pathakCVPR16context} and new stereo and lightfield views~\cite{DeepStereo,LearningViewSynthesis} given a set of input images. Similarly, 3D models can be completed from a partial 3D shape~\cite{dai2017complete,Sung15completion}.
Our problem is different since different glyphs in the same font share the same style, but not structure (unlike one object under different viewpoints).
Various geometry modeling techniques have been proposed for learning structural priors from example 3D shapes~\cite{Huang2015deeplearningsurfaces, Kalogerakis2012ShapeSynthesis} and transferring style from a few examples to an input model~\cite{Lun2016StyleTransfer}.
Font data provides a cleaner factorization of style and content that we leverage in our approach.


\section{Multi-Content GAN Architecture}
We propose an end-to-end network to take a subset of stylized images of specific categories (such as font glyphs) and predict the whole set of stylistically similar images. We have specifically designed our model for the font generation problem to predict the set of letters from A to Z for in-the-wild fonts with a few observed letters. We divide this problem into two parts: glyph generation and texture transfer. Our first network, called GlyphNet, predicts glyph masks while our second network, called OrnaNet, fine-tunes color and ornamentation of the generated glyphs from the first network. Each sub-network follows the conditional generative adversarial network (cGAN) architecture~\cite{isola2016image} modified for its specific purpose of stylizing glyphs or ornamentation prediction. We assume the label for each observed letter is known for the model and thus, skip the need for categorizing each letter into the 26 letters. 
In the following sections, we will first summarize the cGAN model, and then discuss our proposed GlyphNet and OrnaNet architectures and stack them together in an end-to-end final design which we refer to as \textit{MC-GAN}.

\subsection{Conditional Generative Adversarial Networks}

\begin{figure*}
\centering
\includegraphics[width=\textwidth]{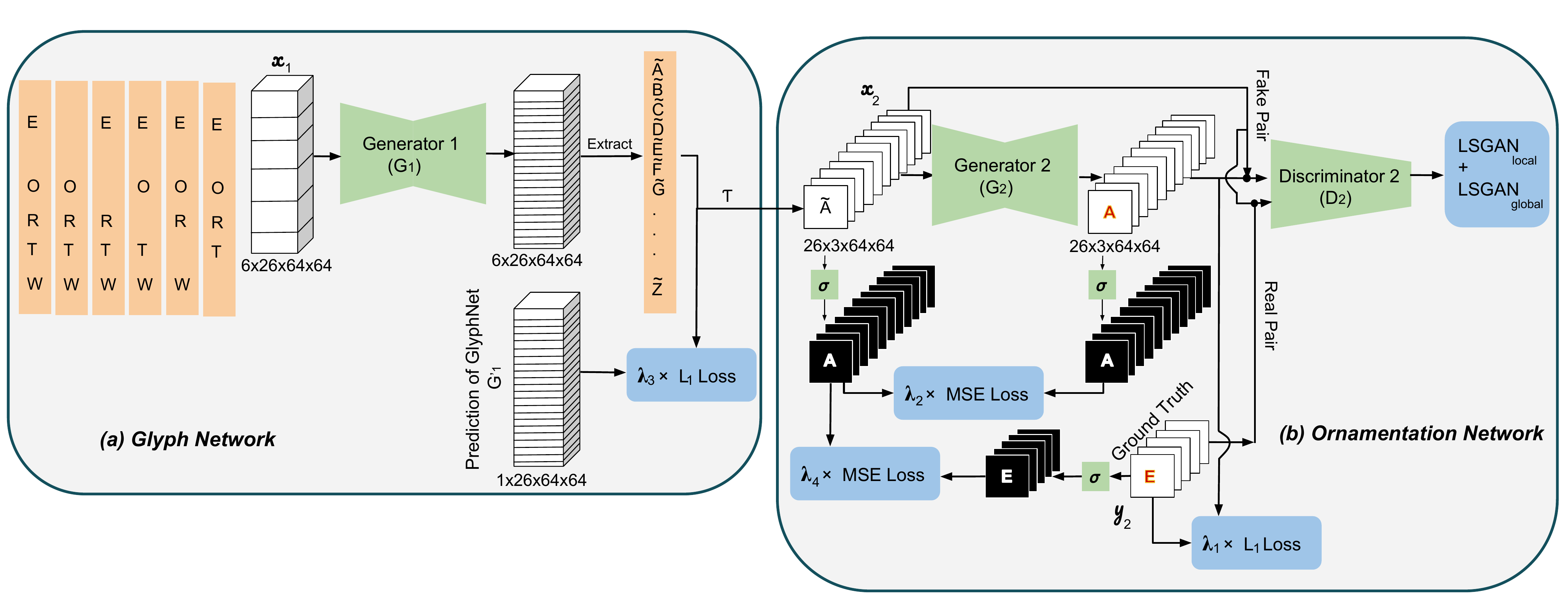}
\caption{Schematic of our end-to-end MC-GAN model including \textbf{(a)} GlyphNet and \textbf{(b)} OrnaNet. Inputs and Outputs are illustrated in white, network layers in green, and loss functions are shown in blue. We use a leave-one-out approach among all observed letters of a word like \textit{TOWER} (in orange) to construct a batch of input image stacks to be fed into $G_1$: For each input stack in the batch, we extract the left out generated glyph. In addition, the remaining 21 glyphs will be generated by feeding in all observed letters together. After a reshape and gray-scale channel repetition, $\mathcal{T}$, these extracted generated glyphs, $\tilde{\text{A}}, \tilde{\text{B}}, \cdots, \tilde{\text{Z}}$ will be fed into OrnaNet.}
\label{fig:diagram}
\end{figure*}

Starting from a random noise vector $z$, generative adversarial networks (GANs)~\cite{goodfellow2014generative} train a model to generate images $y$ following a specific distribution by adversarially training a generator versus a discriminator ($z \rightarrow y$). While the discriminator tries to distinguish between real and fake images, the generator opposes the discriminator by trying to generate realistic-looking images. In the conditional GAN (cGAN) scenario~\cite{isola2016image, mirza2014conditional}, this mapping is modified by feeding an observed image $x$ alongside the random noise vector to the generator ($\{x,z\}\rightarrow y$), and thus,  the adversary between generator and discriminator is formulated as the following loss function:
\begin{eqnarray}
&\mathcal{L}_\text{cGAN} (G,D) = \mathbb{E}_{x,y\sim p_\text{data}(x,y)}[\log D(x,y)]\nonumber\\
&+ \mathbb{E}_{x\sim p_\text{data}(x),z\sim p_z(z)} [1-\log D(x, G(x,z))],
\end{eqnarray} where $G$ and $D$ minimize and maximize this loss function, respectively.

Given the ground truth output of the generator, it is also beneficial to force the model to generate images which are close to their targets through an $L_1$ loss function in addition to fooling the discriminator. The generator's objective can be summarized as:
\begin{eqnarray}
G^* = \arg \min_G \max_D \mathcal{L}_\text{cGAN}(G,D) + \lambda \mathcal{L}_{L_1}(G),
\end{eqnarray} where $\mathcal{L}_{L_1}(G) = \mathbb{E}_{x,y\sim p_\text{data}(x,y), z\sim p_z(z)} [\|y-G(x,z)\|_1]$.

We follow this conditional GAN setting in each of our sub-networks to generate the whole set of letters with a consistent style, $y$, by observing only a few examples fed in as a stack, $x$.   Similar to ~\cite{isola2016image}, we ignore random noise as the input to the generator, and dropout is the only source of randomness in the network.

\subsection{Glyph Network}
\label{sec:glyph}
Generalizing all 26 capital letters of a font from a few example glyphs requires capturing correlations and similarities among source letters and the unseen ones. Our GlyphNet learns such correlations automatically in order to generate a whole set of stylistically similar glyphs. We study this behavior in Section~\ref{sec:corr}.

Due to the style similarity among all content images, we add one input channel for each individual glyph in our proposed GlyphNet resulting in a ``glyph stack'' in both input and the generated output (as illustrated in Figure~\ref{fig:glyph}). A basic  tiling of all 26 glyphs into a single image, however, fails to capture correlations among them specifically for those far from each other along the image length. This occurs due to the smaller size of convolution receptive fields than the image length within a reasonable number of convolutional layers. 

With our novel input glyph stack design, correlation between different glyphs are learned across network channels in order to transfer their style automatically. We employ our generator, $G_1$, based on the image transformation network introduced in~\cite{johnson2016perceptual} including six ResNet blocks. The full architectual specification of both GlyphNet and OrnaNet are provided in Appendix~\ref{sec:net-arch}.

We consider $64\times 64$ glyphs in gray-scale resulting in the input and output dimension of $B\times 26\times 64 \times 64$ for the 26 capital English alphabets, with $B$ indicating batch size. Following the PatchGAN model proposed by~\cite{isola2016image}, we apply a $21\times 21$ local discriminator with three convolutional layers on top of the generated output stack in order to discriminate between real and fake local patches resulting in a receptive field size equal to 21. In parallel, we add two extra convolutional layers as a global discriminator, resulting in a receptive field covering the whole image to distinguish between realistic font images and generated ones. In Figure~\ref{fig:glyph}, our local and global discriminators are shown within one discriminator block and will be referred as $D_1$.

For higher quality results and to stabilize GAN training~\cite{zhu2017unpaired}, we use two least squares GAN (LSGAN) loss functions~\cite{mao2016least} on our local and global discriminators added with an $L_1$ loss penalizing deviation of generated images $G_1(x_1)$ from their ground truth $y_1$:
\begin{eqnarray}
\mathcal{L}(G_1) &=& \lambda \mathcal{L}_{L_1}(G_1) + \mathcal{L}_\text{LSGAN} (G_1,D_1)\nonumber \\
 &=& \lambda \mathbb{E}_{x_1,y_1\sim p_\text{data}(x_1,y_1)}[\|y_1-G_1(x_1)\|_1] \\
 && +\mathbb{E}_{y_1 \sim p_\text{data}(y_1)}  [(D_1(y_1)-1)^2] \nonumber\\
 && + \mathbb{E}_{x_1\sim p_\text{data}(x_1)}[D_1(G_1(x_1))^2],\nonumber
\end{eqnarray} where 
\begin{eqnarray}
\mathcal{L}_\text{LSGAN} (G_1,D_1) = \mathcal{L}_\text{LSGAN}^{\text{local}} (G_1,D_1) + \mathcal{L}_\text{LSGAN}^{\text{global}} (G_1,D_1). \nonumber
\end{eqnarray}
We train this network on our collected $10 \mathrm{K}$ font data set (Section~\ref{sec:dataset}) where in each training iteration, $x_1$ includes a randomly chosen subset of $y_1$ glyphs with the remaining input channels being zeroed out. We will refer to this \textit{trained model} as $G'_1$ in the following sections. We explored adding in a separate input indicator channel denoting which of the glyphs are present, but did not find this to significantly affect the quality of the generator.

While we pre-train the GlyphNet using the conditional discriminator, we will remove this discriminator when training the joint network (Section~\ref{sec:end2end}).

\subsection{Ornamentation Network}
Our second sub-network, OrnaNet, is designed to transfer ornamentation of the few observed letters to the gray-scale glyphs through another conditional GAN network consisting of a generator, $G_2$, and a discriminator, $D_2$. Feeding in the glyphs as input images, $x_2$, this network generates outputs, $G_2(x_2)$, enriched with desirable color and ornamentation. The main difference between our proposed OrnaNet and GlyphNet lies in the dimension and type of inputs and outputs, as well as in how broad vs. specific the model is in generating images with a particular style; the generator and conditional discriminator architecture is otherwise identical to GlyphNet.

While GlyphNet is designed to generalize glyph correlations across all our training fonts, OrnaNet is specialized to apply only the specific ornamentation observed in a given observed font. It is trained only on the small number of observations available. Moreover, inputs and outputs of the OrnaNet include a batch of images with three RGB channels (similar to the format of the input and output images used in~\cite{isola2016image}) where the the input channels are repeats of the gray-scale glyphs. In the next section, we will describe how to combine our GlyphNet and OrnaNet in an end-to-end manner in order to generate stylized glyphs in an ornamented typeface.
 
\subsection{End-to-End Network}
\label{sec:end2end}
 The goal of our end-to-end model, illustrated in Figure~\ref{fig:diagram}, is to generalize both style and ornamentation of the observed letters to the unobserved ones. For this purpose, we generate all 26 glyphs including the observed ones through the pre-trained GlyphNet and feed them to the OrnaNet (initialized with random weights) to be fine-tuned. To accomplish this, we use a leave-one-out approach to cycle all possible unobserved letters:
 
 For instance, given 5 observed letters of the word \textit{TOWER} shown in Figure~\ref{fig:diagram}, we first use 4 letters \textit{T, O, W, E} as the given channels in a $1\times 26 \times 64\times 64$ input stack and feed it to the pre-trained GlyphNet to generate all 26 letters and then extract the one fake glyph, \textit{$\tilde{\text{R}}$}, not included in the input stack. Repeating this process would generate all of the 5 observed letters from the pre-trained GlyphNet. Similarly, we extract the 21 remaining letters from the pre-trained model by feeding in a $1\times 26 \times 64\times 64$ input stack filled with all 5 observed letters simultaneously while zeroing out all other channels. This whole process can be summarized by passing 6 input stacks each with dimension $1\times 26 \times 64\times 64$ through GlyphNet as a batch, extracting the relevant channel from each output stack, and finally concatenating them into one $1\times 26 \times 64\times 64$ output. After a reshape transformation and gray-scale channel repetition, represented by $\mathcal{T}$, we can transform this generated output to 26 images with dimension $ 3\times 64 \times 64$ and feed them as a batch, $x_2$, to OrnaNet. This leave-one-out approach enables OrnaNet to generate high quality stylized letters from coarse generated glyphs.

To stabilize adversarial training of the OrnaNet generator ($G_2$) and discriminator ($D_2$), we likewise use an LSGAN loss added with an $L_1$ loss function on generated images of the observed letters, $x_2$, and their ground truth, $y_2$. Moreover, to generate a set of color images with clean outlines, we minimize the mean square error (MSE) between  binary masks of the outputs and inputs of the generator in OrnaNet which are fake color letters, $G_2(x_2)$, and fake gray-scale glyphs, $x_2$, respectively. Binary masks are obtained by passing images through a sigmoid function, indicated as $\sigma$ in~\eqref{eq:orna}. In summary, the loss function applied on top of the OrnaNet in the end-to-end scenario can be written as:
\begin{eqnarray}
\label{eq:orna}
\mathcal{L}(G_2) &=& \mathcal{L}_\text{LSGAN} (G_2,D_2) + \lambda_1 \mathcal{L}_{L_1}(G_2) + \lambda_2 \mathcal{L}_\text{MSE} (G_2) \nonumber \\
 &=& \mathbb{E}_{y_2 \sim p_\text{data}(y_2)}  [(D_2(y_2)-1)^2] \nonumber\\
 && + \mathbb{E}_{x_2\sim p_\text{data}(x_2)}[D_2(G_2(x_2))^2]\\
 && + \mathbb{E}_{x_2,y_2\sim p_\text{data}(x_2,y_2)}\big[\lambda_1  \|y_2-G_2(x_2)\|_1 \nonumber \\
 && + \lambda_2  (\sigma(y_2)-\sigma(G_2(x_2)))^2\big], \nonumber
\end{eqnarray} where $x_2 = \mathcal{T}(G_1(x_1))$ and 
\begin{eqnarray}
\mathcal{L}_\text{LSGAN} (G_2,D_2) = \mathcal{L}_\text{LSGAN}^{\text{local}} (G_2,D_2) + \mathcal{L}_\text{LSGAN}^{\text{global}} (G_2,D_2). \nonumber
\end{eqnarray}.

In the final end-to-end training, we do not use discriminator $D_1$ in the GlyphNet and instead, OrnaNet plays the role of a loss function by back propagating the gradients of the objective in~\eqref{eq:orna} to improve style of the generated glyphs. Adding a weighted $L_1$ loss on top of the generator in GlyphNet, $G_1$, also penalizes deviating from the predictions of the pre-trained GlyphNet, $G'_1$.
We also add an MSE loss function between binary masks of fake versions of the observed glyphs, $\mathcal{T}(G_1(x_1))$, and masks of their corresponding ground truth glyphs, $y_2$. Putting this all together, the gradients of the following loss functions would be passed through GlyphNet in addition to the gradient coming from OrnaNet:
\begin{eqnarray}
\label{eq:glyph-end}
\mathcal{L}(G_1) &=& \lambda_3 \mathcal{L}_{w,L_1}(G_1) + \lambda_4 \mathcal{L}_\text{MSE}(G_1)\nonumber\\
&=& \mathbb{E}_{x_1\sim p_\text{data}(x_1),y_2\sim p_\text{data}(y_2)} \big[ \nonumber\\
&&\lambda_3 \sum_{i=1}^{26} w_i\times|G_1^i(x_1) - G_1^{'i}(x_1)| +\nonumber\\
&&\lambda_4 (\sigma(y_2) - \sigma(\mathcal{T}(G_1(x_1)))^2 \big],
\end{eqnarray} where $w_i$ allows us to apply different weights to observed vs. unobserved glyphs. Ratio between different terms in loss functions in ~\eqref{eq:orna},~\eqref{eq:glyph-end} is defined based on hyper-parameters $\lambda_1$ to $\lambda_4$. Moreover, as mentioned in Section~\ref{sec:glyph}, $G'_1(x)$ indicates the prediction of the pre-trained GlyphNet before being updated through end-to-end training. 

\begin{figure}[t!]
\centering
\includegraphics[width=\linewidth]{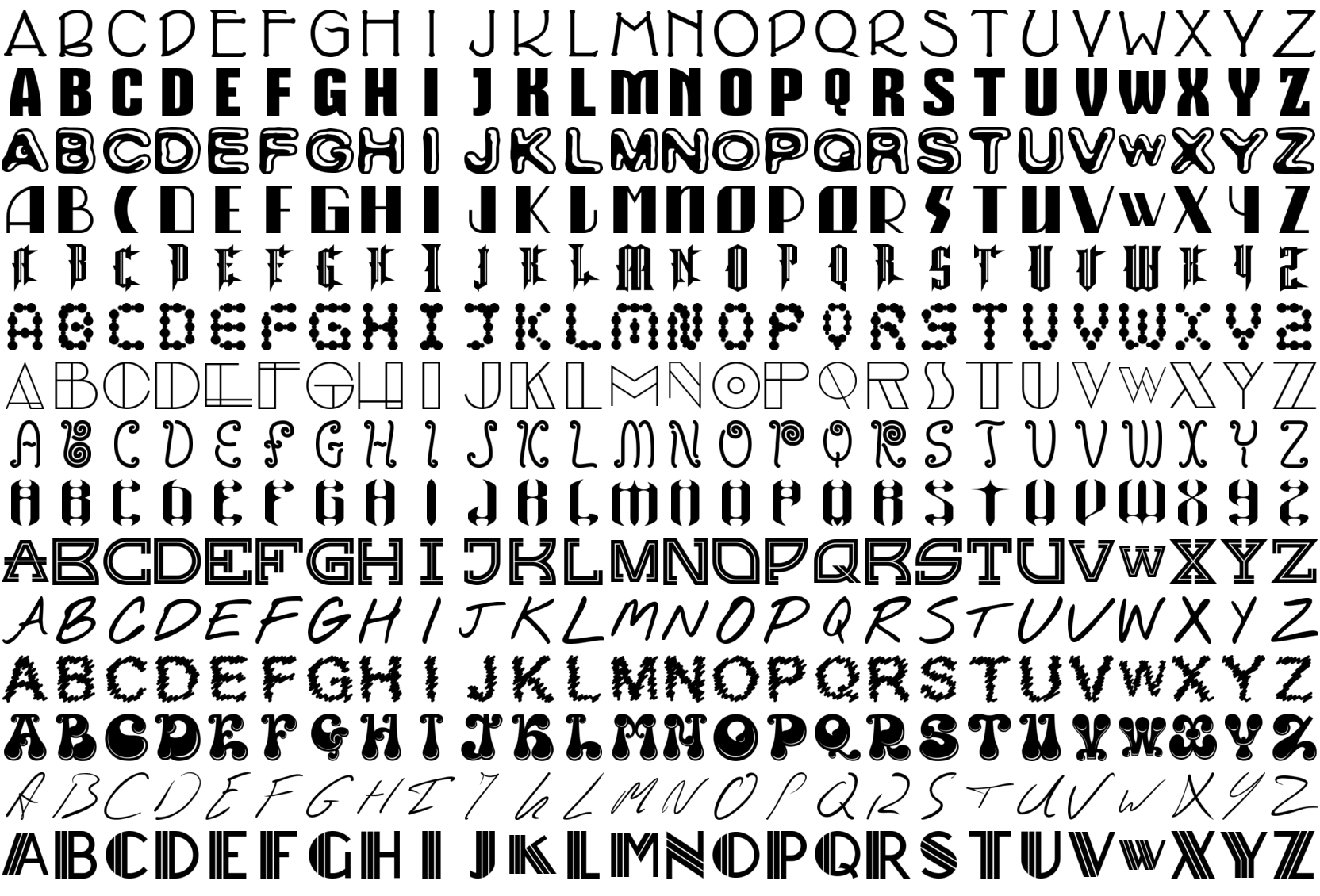}
\caption{Random subset of our $10\mathrm{K}$ gray-scale font dataset}
\label{fig:data}
\end{figure}

\begin{figure}[t!]
\centering
\includegraphics[width=\linewidth]{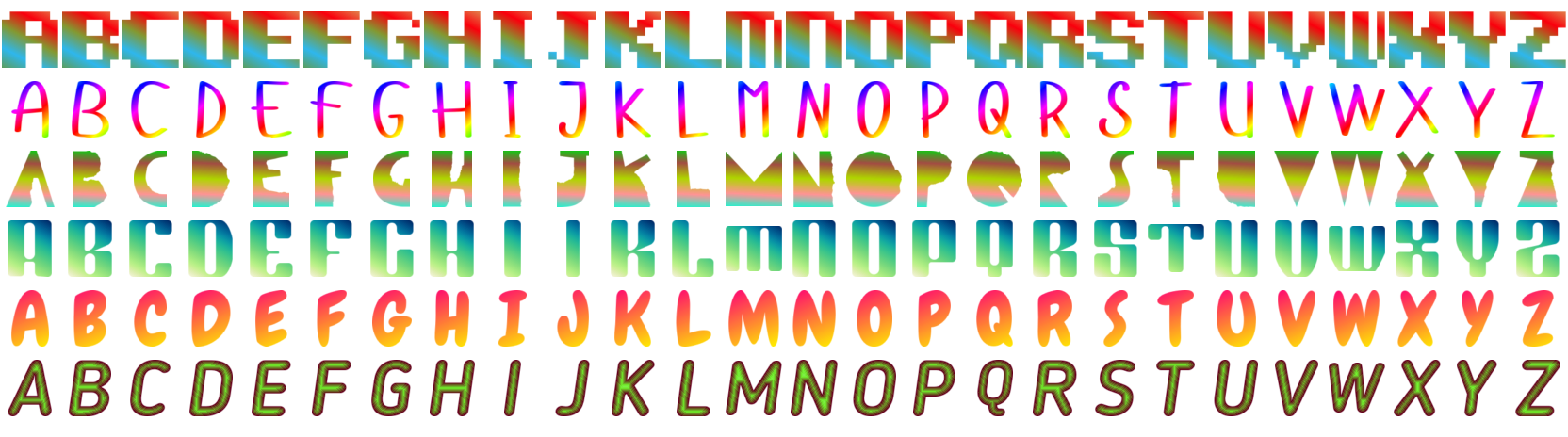}
\caption{Example synthetic color gradient fonts}
\label{fig:data_color}
\vspace{-1em}
\end{figure}

\section{Font Dataset}
\label{sec:dataset}
We have collected a dataset including $10\mathrm{K}$ gray-scale Latin fonts each with 26 capital letters. We process the dataset by finding a bounding box around each glyph and resize it so that the larger dimension reaches 64 pixels, then pad to create $64\times 64$ glyphs. A few exemplar fonts from our dataset are depicted in Figure~\ref{fig:data}. These fonts contain rich information about inter-letter correlations in font styles, but only encode glyph outlines and not font ornamentations.
To create a baseline dataset of ornamented fonts, we apply random color gradients and outlining on the gray-scale glyphs, two random color gradients on each font, resulting in a $20\mathrm{K}$ color font data set. A few examples are shown in Figure~\ref{fig:data_color}. Size of this data set can be arbitrarily increased through generating more random colors. These gradient fonts do not have the same distribution as in-the-wild ornamentations but can be used for applications such as network pre-training.

\section{Experiments and Results}
\label{sec:experiments}
We demonstrate the quality of our end-to-end model predictions on multiple fonts with various styles and decorations. First, we study the advantage of various components of our model through different ablation studies. Next, we will show the significant improvement obtained by our model in transferring ornamentations on our synthesized glyphs compared with patch-based text effect transfer approach~\cite{yang2016awesome}. In the following experiments, we have set $\lambda_1=300, \lambda_2=300$ if $\text{epoch}< 200 \text{ and } \lambda_2=3$ otherwise, $\lambda_3=10, \lambda_4=300$, while $w_i=10$ if $i$ is an observed glyph and $w_i=1$ otherwise.

\begin{figure*}
\centering
\includegraphics[width=\textwidth]{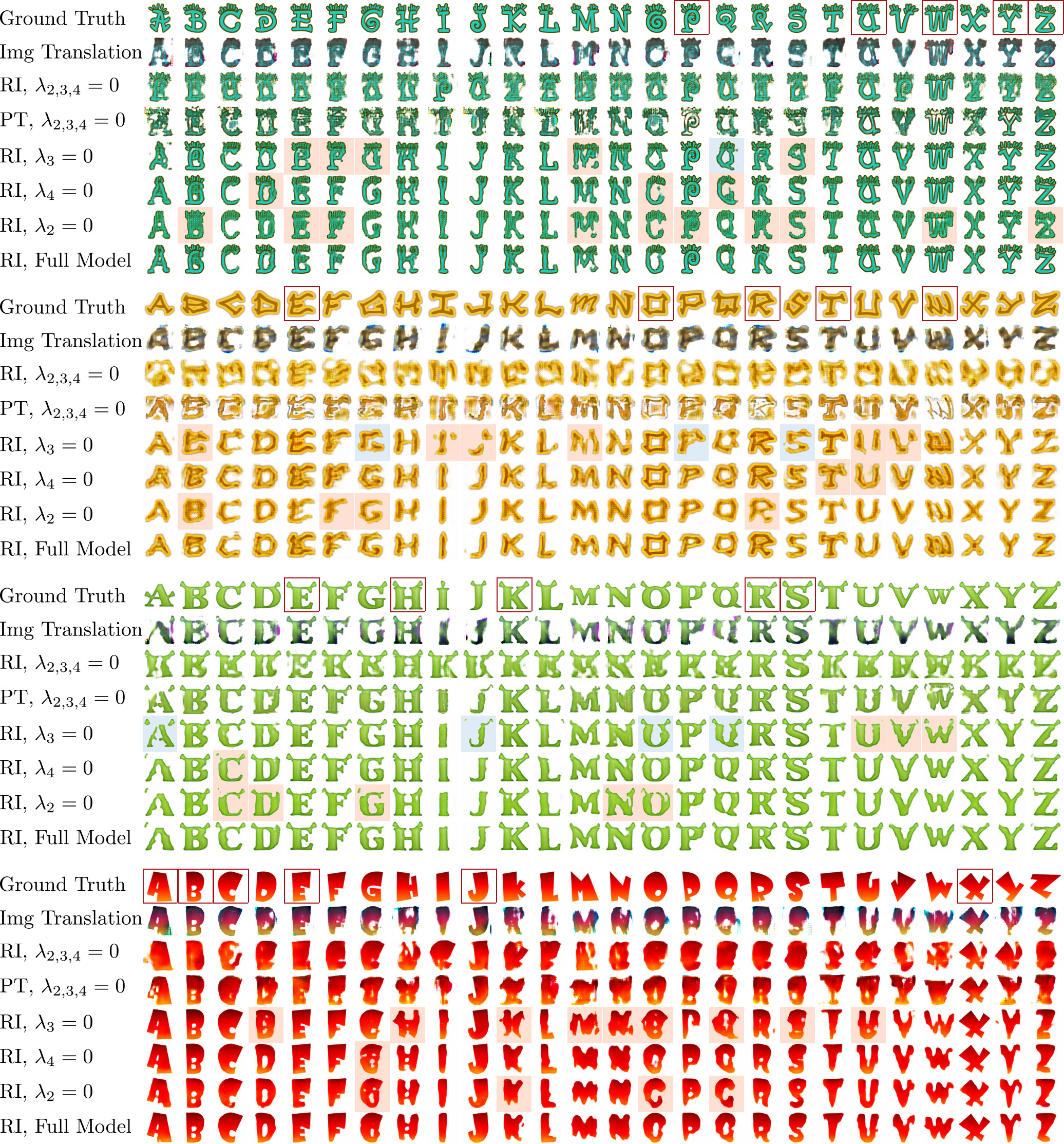}
\caption{Ablation study on our MC-GAN model components: For each exemplar font, we show ground truth (\textbf{1st row}), observed letters (red squares in the 1st row), predictions of a baseline image translation network (\textbf{2nd row}), predictions of our end-to-end model with randomly initialized (RI) OrnaNet and $\lambda_2=\lambda_3=\lambda_4=0$ (\textbf{3rd row}), with pre-trained (PT) OrnaNet weights and $\lambda_2=\lambda_3=\lambda_4=0$ (\textbf{4th row}), selectively disabled loss terms (\textbf{rows 5-7}), and the full end-to-end MC-GAN model (\textbf{bottom row}). Style transfer improvements by $\lambda_3$ are highlighted in blue and degradation in the predictions by omitting each individual regularizer is highlighted in red.} 
\label{fig:ablation}
\end{figure*}

  For evaluation, we download ornamented fonts from the web\footnote[1]{http://www6.flamingtext.com/All-Logos}.  For all experiments in sections~\ref{sec:image-trans} and~\ref{sec:ablation}, we have made sure that all font examples used in these studies were not included in our $10\mathrm{K}$ font training set by manually inspecting nearest neighbors computed over the black-and-white glyphs.

\subsection{Image Translation Baseline}
\label{sec:image-trans}
To illustrate the significant quality improvement of our end-to-end approach, we have implemented a baseline image-to-image translation network~\cite{isola2016image} for this task. In this baseline approach, we consider channel-wise letters in input and output stacks with dimensions $B\times 78\times 64 \times 64$, where $B$ stands for training batch size and $78$ corresponds to the 26 RGB channels. The input stack is given with ``observed'' color letters while all letters are generated in the output stack. We train this network on our color font data set where we have applied randomly chosen color gradients on each gray-scale font. Feeding in a random subset of RGB letters of an arbitrary font into this model during test time, it is expected to generate stylistically similar 26 letters. Results of this model are shown in the second rows of Figure~\ref{fig:ablation} for each example font. Observe that while the network has learned rough aspects of glyph structure, the predictions do not follow a consistent color or ornamentation scheme, as the network is not able to effectively specialize for the provided ornamentation style. Similar artifacts are observed even when evaluating on a test set derived from our simplified color-gradient dataset (see Section~\ref{sec:color}).

\subsection{Ablation Study}
\label{sec:ablation}
In Figure~\ref{fig:ablation} we demonstrate the incremental improvement of our proposed regularizers, $\mathcal{L}_{w,L_1}(G_1), \mathcal{L}_{MSE}(G_1)$, and $\mathcal{L}_{MSE}(G_2)$. We found that pre-training on our OrnaNet on gradient-based ornamentations was not helpful, and that the best result comes from a random initialization of OrnaNet and using all the proposed loss terms.

As mentioned in Section~\ref{sec:end2end}, $\mathcal{L}_{w,L_1}(G_1)$ prevents network predictions from going far from the original pre-trained predictions of the GlyphNet. However, it also reduces the freedom in modifying the style of the new glyphs during the end-to-end training. We show this trade-off in the fourth rows of each instance font in Figure~\ref{fig:ablation} by highlighting letters with additional artifacts in red and improved letters in blue when this regularizer is excluded from the network. The other two MSE loss regularizers weighted by $\lambda_2$ and $\lambda_4$ prevent blurry predictions or noisy artifacts to appear on the generated gray-scale and color letters.

\begin{figure}
\centering
\includegraphics[width=0.5\textwidth]{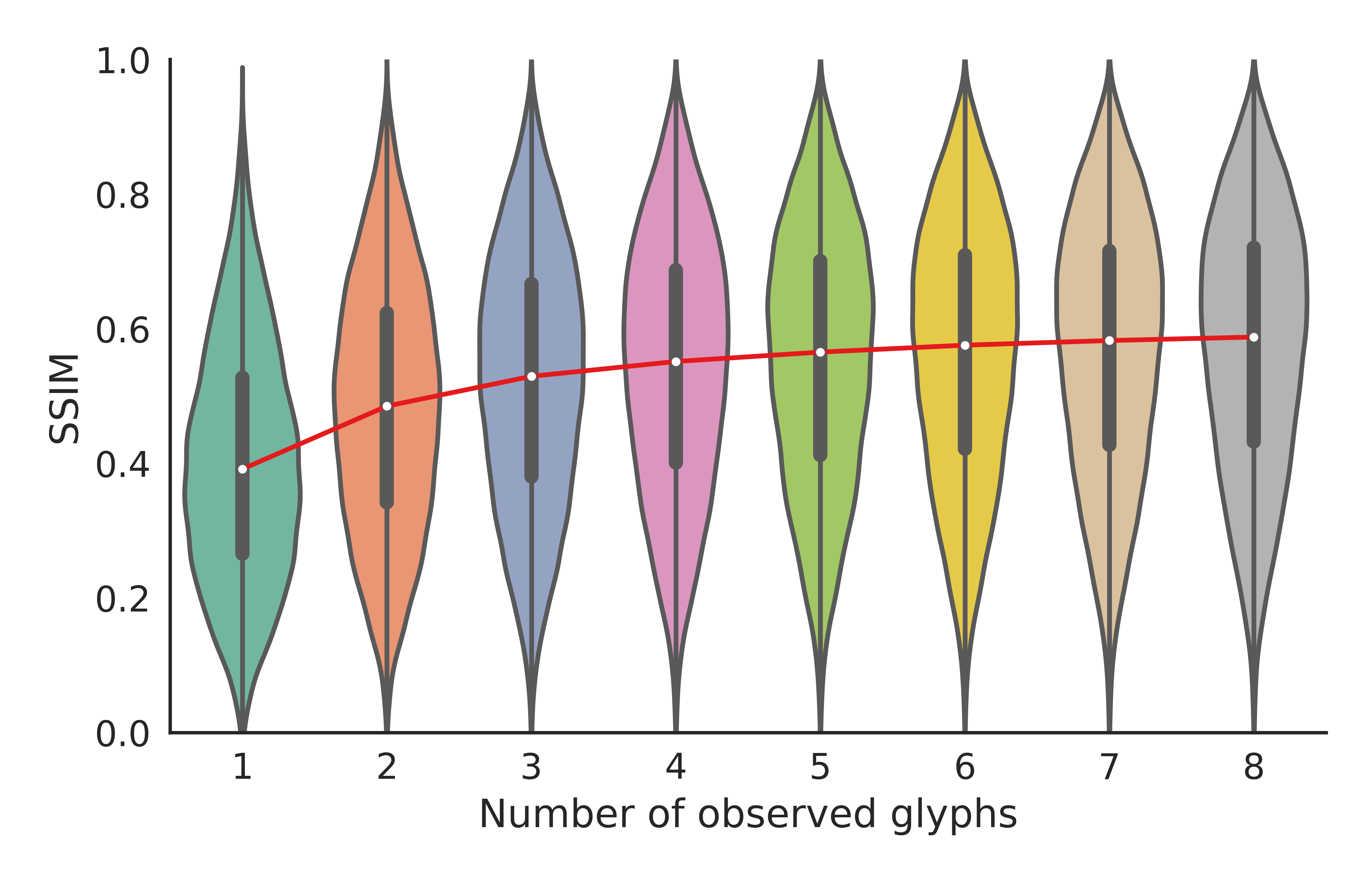}
\caption{Effect of number of observed glyphs on the quality of GlyphNet predictions. Red line is passing through median of each distribution.} 
\label{fig:num-given}
\end{figure}

\subsection{Automatic Learning of Correlations between Contents}
\label{sec:corr}
Automatic learning of the correlations existing between different letters is a key factor in transferring style of the few observed letters in our multi-content GAN. In this section, we study such correlations through the structural similarity (SSIM) metric on a random subset of our $10\mathrm{K}$ font data set consisting of 1500 examples. For each instance, we randomly keep one of the 26 glyphs and generate the rest through our pre-trained GlyphNet. 

\begin{figure*}[hpbt!]
\centering
\includegraphics[width=0.85\textwidth]{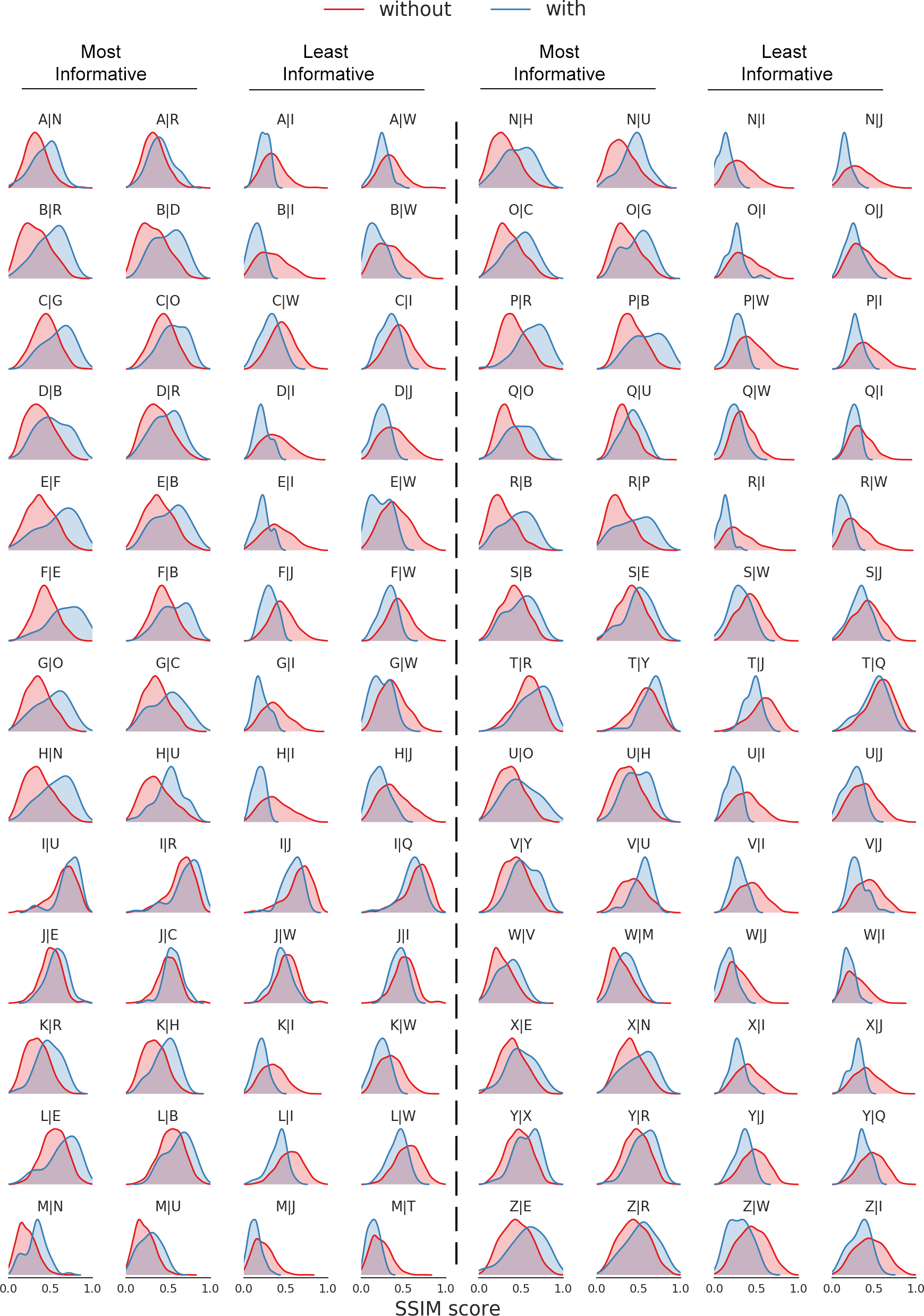}
\caption{Distributions ($\alpha$|$\beta$) over SSIM scores for generating letter $\alpha$ given $\beta$ in blue and given any other letter rather than $\beta$ in red. Distributions for the most informative given letters $\beta$ in generating each glyph $\alpha$ is shown in the left of each column while the least informative givens are presented in the right.} 
\label{fig:corr}
\end{figure*}

Computing the structural similarity between each generated glyph and its ground truth, we find 25 distributions over its SSIM scores when a single letter has been observed at a time. In Figure~\ref{fig:corr}, we illustrate the distributions $\alpha|\beta$ of generating letter $\alpha$ when letter $\beta$ is observed (in blue) vs when any other letter rather than $\beta$ is given (in red). Distributions for the two most informative given letters and the two least informative ones in generating each of the 26 letters are shown in this figure. For instance, looking at the fifth row of the figure, letters \textit{F} and \textit{B} are the most constructive in generating letter \textit{E} compared with other letters while \textit{I} and \textit{W} are the least informative ones. As other examples, \textit{O} and \textit{C} are the most guiding letters for constructing \textit{G} as well as \textit{R} and \textit{B} for generating \textit{P}. 
\subsection{Number of Observed Letters}
Here, we investigate the dependency of quality of GlyphNet predictions on the number of observed letters. Similar to Section~\ref{sec:corr}, we use a random subset of our font data set with 1500 example fonts and for each font, we generate 26 letters given $n$ observed ones from our pre-trained GlyphNet. The impact of changing $n$ from 1 to 8 on the distribution of SSIM scores between each unobserved letter and its ground truth is shown in Figure~\ref{fig:num-given}. The slope of the red line passing through the median of each distribution is decreasing as $n$ increases and reaches to a stable point once the number of observations for each font is close to 6. This study confirms the advantage of our multi-content GAN method in transferring style when we have very few examples. 

\subsection{Perceptual Evaluation}
\label{sec:perceptualEvaluation}

\begin{figure*}[hpbt!]
\centering
\includegraphics[width=\textwidth]{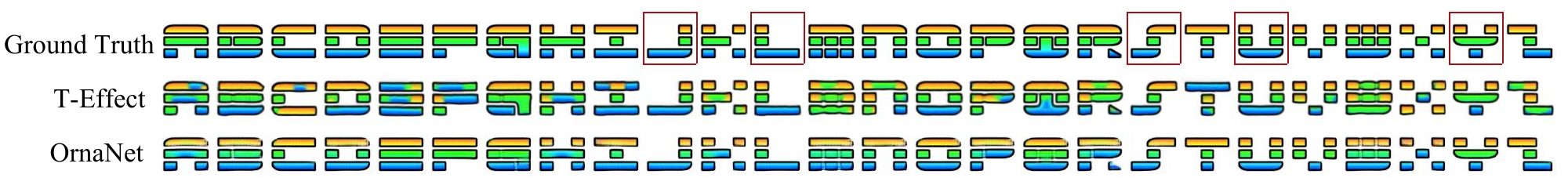}
\caption{Text Effect Transfer~\cite{yang2016awesome} failure example on clean input glyphs.} 
\label{fig:success}
\end{figure*}

\begin{figure*}[hpbt!]
\centering
\includegraphics[width=\textwidth]{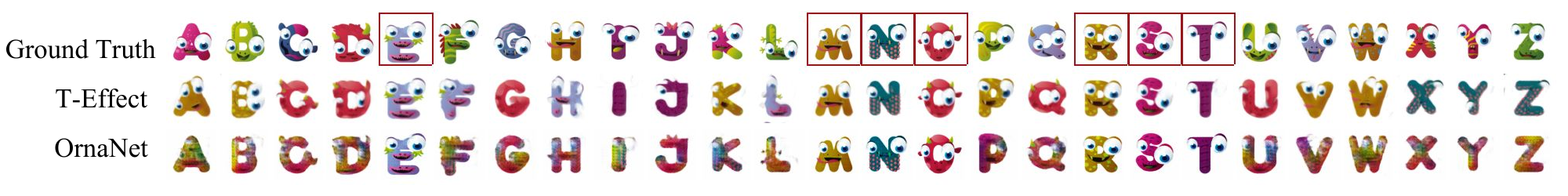}
\caption{Failure cases on clean input glyphs.} 
\label{fig:fail}
\end{figure*}

To evaluate the performance of our model, we compare the generated letters of our end-to-end multi-content GAN against the output of the patch-based synthesis method in~\cite{yang2016awesome}. Since this model is designed only for transferring text decorations on clean glyphs, it is not fully comparable with our approach which synthesizes unobserved letters. To explore this method, we use the predictions of our pre-trained GlyphNet as the input to this algorithm. Moreover, this model transfers stylization from only one input decorated glyph, while our method uses all observed examples simultaneously. Therefore, to enable a fair comparison in transferring ornamentations, we allow their model to choose the most similar glyph among the observed instances to the generated glyph mask using a simple image-space distance metric.

\begin{figure*}
\centering
\includegraphics[width=\textwidth]{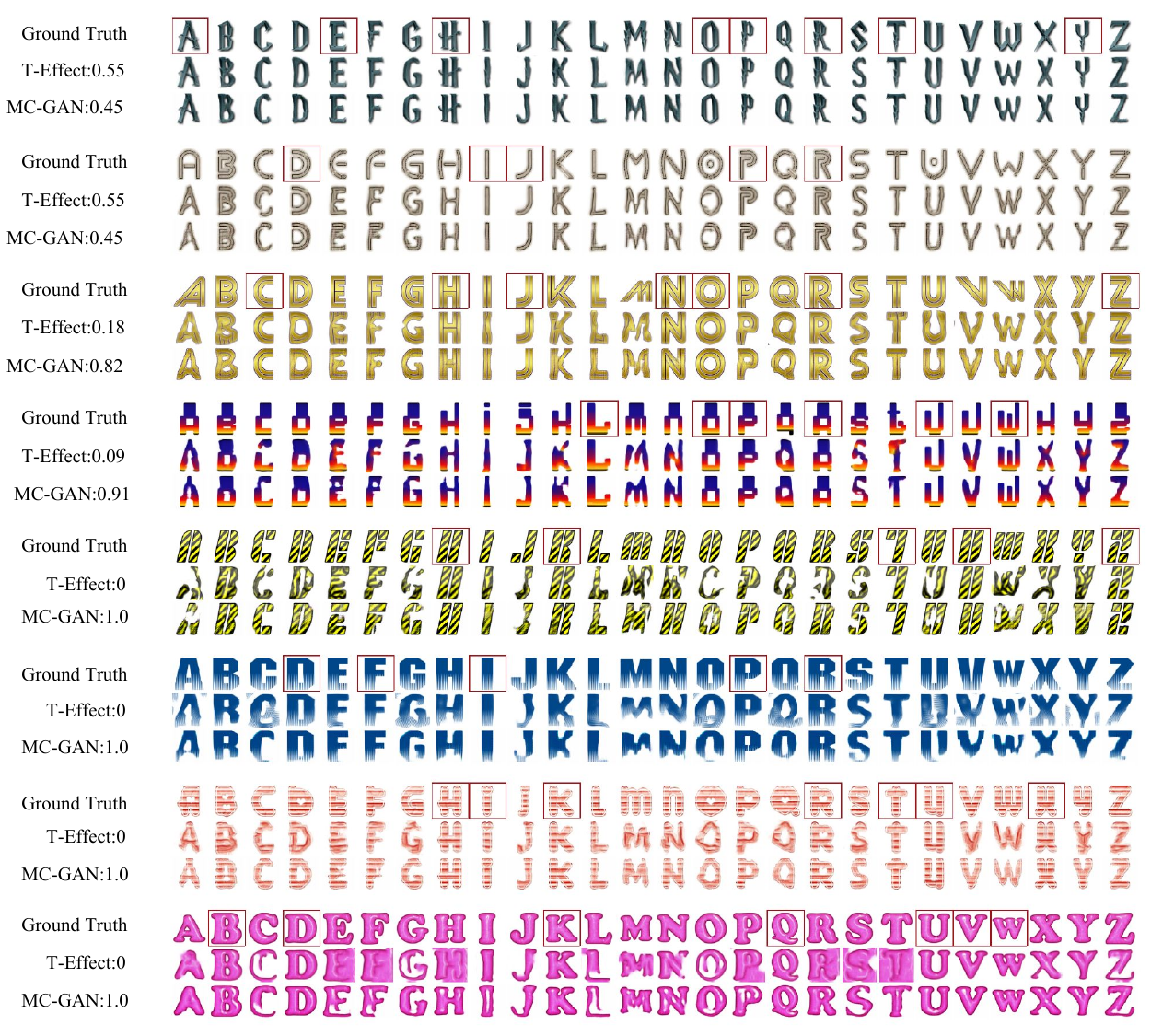}
\caption{Comparison of our end-to-end MC-GAN model (\textbf{3rd rows}) with the text effect transfer approach~\cite{yang2016awesome} using GlyphNet synthesized glyphs (\textbf{2nd rows}). Ground truth glyphs and the observed subset are illustrated in the \textbf{1st row} of each example font. Scores next to each example reveal the percentage of people who preferred the given results.}
\label{fig:comp}
\vspace{12em}
\end{figure*}

\begin{figure*}
\centering
\includegraphics[width=\textwidth]{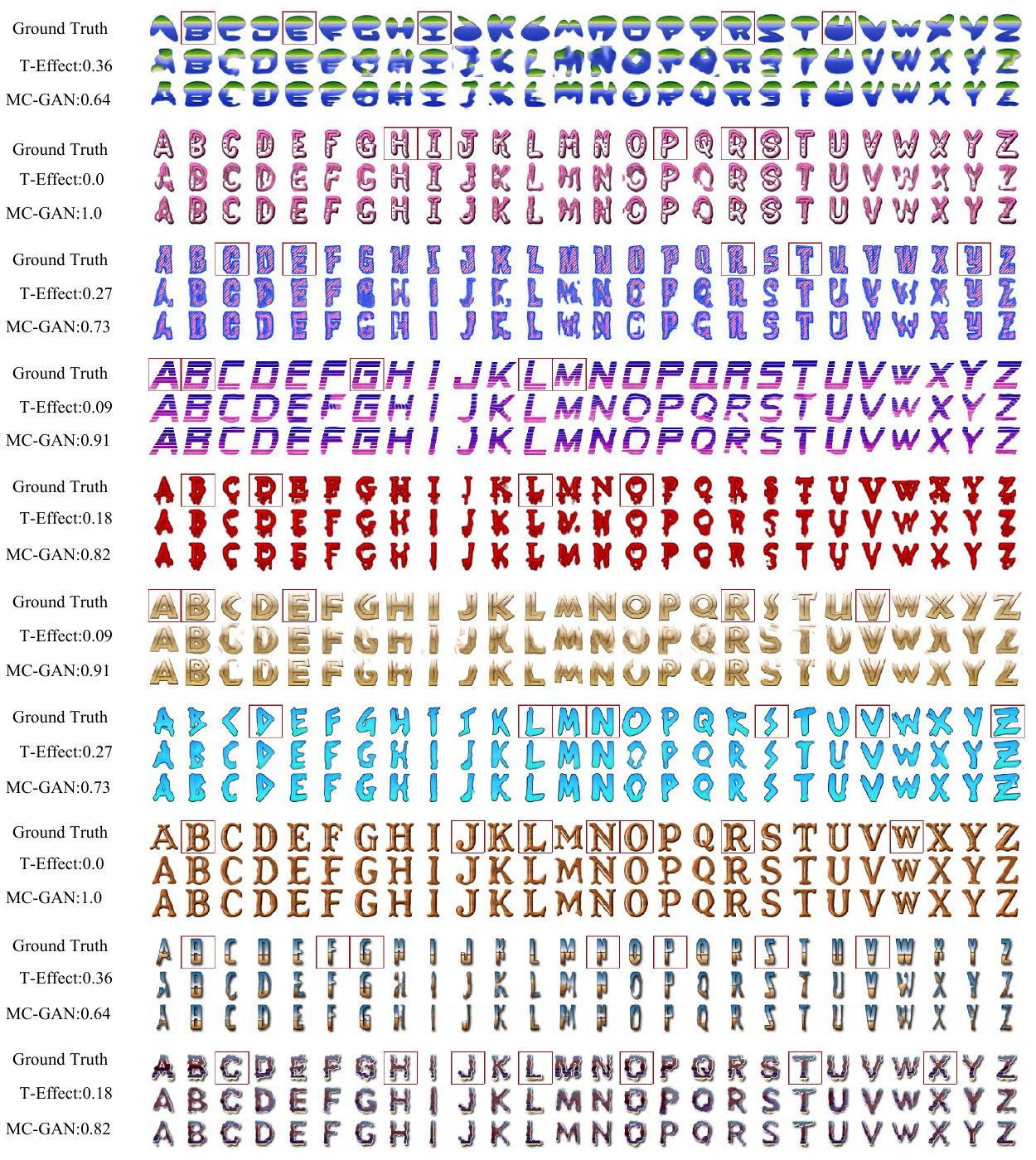}
\caption{Continue - Comparison of our end-to-end MC-GAN model (\textbf{3rd rows}) with the text effect transfer approach~\cite{yang2016awesome} using GlyphNet synthesized glyphs (\textbf{2nd rows}). Ground truth glyphs and the observed subset are illustrated in the \textbf{1st row} of each example font. Scores next to each example reveal the percentage of people who preferred the given results.}
\label{fig:comp-p2}
\vspace{4em}
\end{figure*}

\begin{figure*}
\centering
\includegraphics[width=\textwidth]{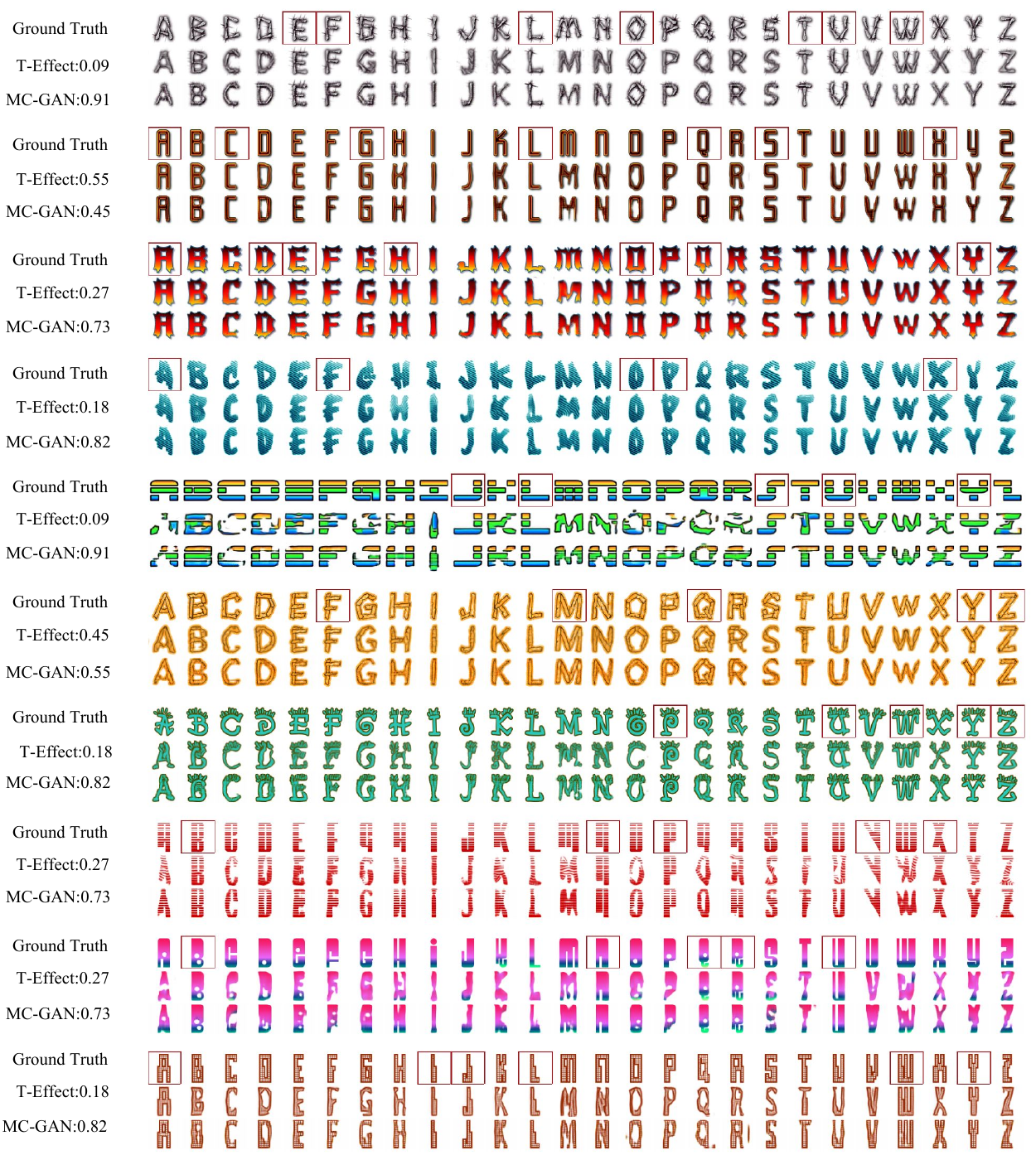}
\caption{Continue - Comparison of our end-to-end MC-GAN model (\textbf{3rd rows}) with the text effect transfer approach~\cite{yang2016awesome} using GlyphNet synthesized glyphs (\textbf{2nd rows}). Ground truth glyphs and the observed subset are illustrated in the \textbf{1st row} of each example font. Scores next to each example reveal the percentage of people who preferred the given results.}
\label{fig:comp-p3}
\vspace{4em}
\end{figure*}

\begin{figure*}
\centering
\includegraphics[width=\textwidth]{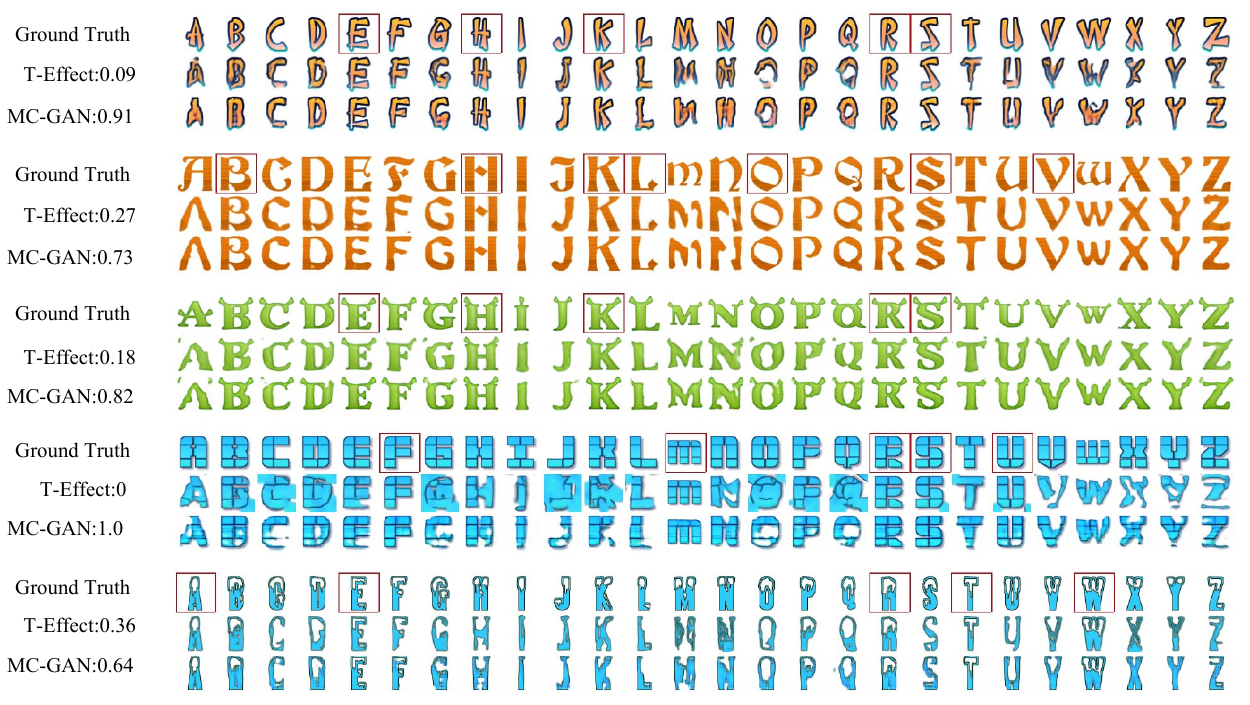}
\caption{Continue - Comparison of our end-to-end MC-GAN model (\textbf{3rd rows}) with the text effect transfer approach~\cite{yang2016awesome} using GlyphNet synthesized glyphs (\textbf{2nd rows}). Ground truth glyphs and the observed subset are illustrated in the \textbf{1st row} of each example font. Scores next to each example reveal the percentage of people who preferred the given results.}
\label{fig:comp-p4}
\vspace{30em}
\end{figure*}

We generated the output of both methods on 33 font examples downloaded from the web and asked 11 people to choose which character set they preferred when presented with the observed letters and the full glyph results of both methods. Overall users preferred the result of our method 80.0\% of the time. We visualize these examples in Figures~\ref{fig:comp},~\ref{fig:comp-p2},~\ref{fig:comp-p3},~\ref{fig:comp-p4} including ground truth and given letters (first rows), predictions of the text effect transfer method~\cite{yang2016awesome}which are applied on top of the glyphs synthesized by our GlyphNet (second rows), and predictions of our full end-to-end model in the last rows. The two best and two worst scoring results for each method are shown on the top and bottom examples of Figure~\ref{fig:comp}. 

The text effect transfer approach is designed to generate text patterns on clean glyphs but mostly fails to transfer style given our synthesized gray-scale letters. In addition, due to their dependency on a patch matching based algorithm, they often cannot transfer style correctly when the shape of the given and new letters are not very similar (e.g., they cannot transfer straight line patterns when there is a curvature in their new input glyph as clear from the sixth and seventh examples in Figure~\ref{fig:comp}).

\subsection{Ground Truth Glyph Ornamentation}

We further compare the performance of our ornamentation network against patch-based synthesis in the case where we are given correct grayscale glyphs (i.e. the ground-truth for GlyphNet). Figure~\ref{fig:success} indicates a failure mode of patch-based effect transfer, where spatial patterns present in the input are often violated. Figure~\ref{fig:fail} represents a failure mode of both methods: our method averages over the distinct colors present and does not always generate the observed ornamentation such as eyes, while patch-based effect transfer better preserves the input color distrubtion but can still fail to capture the frequency of stylistic elements.

\subsection{Generalization on Synthetic Color Font Dataset}
\label{sec:color}
\begin{figure*}
\centering
\includegraphics[width=0.95\textwidth]{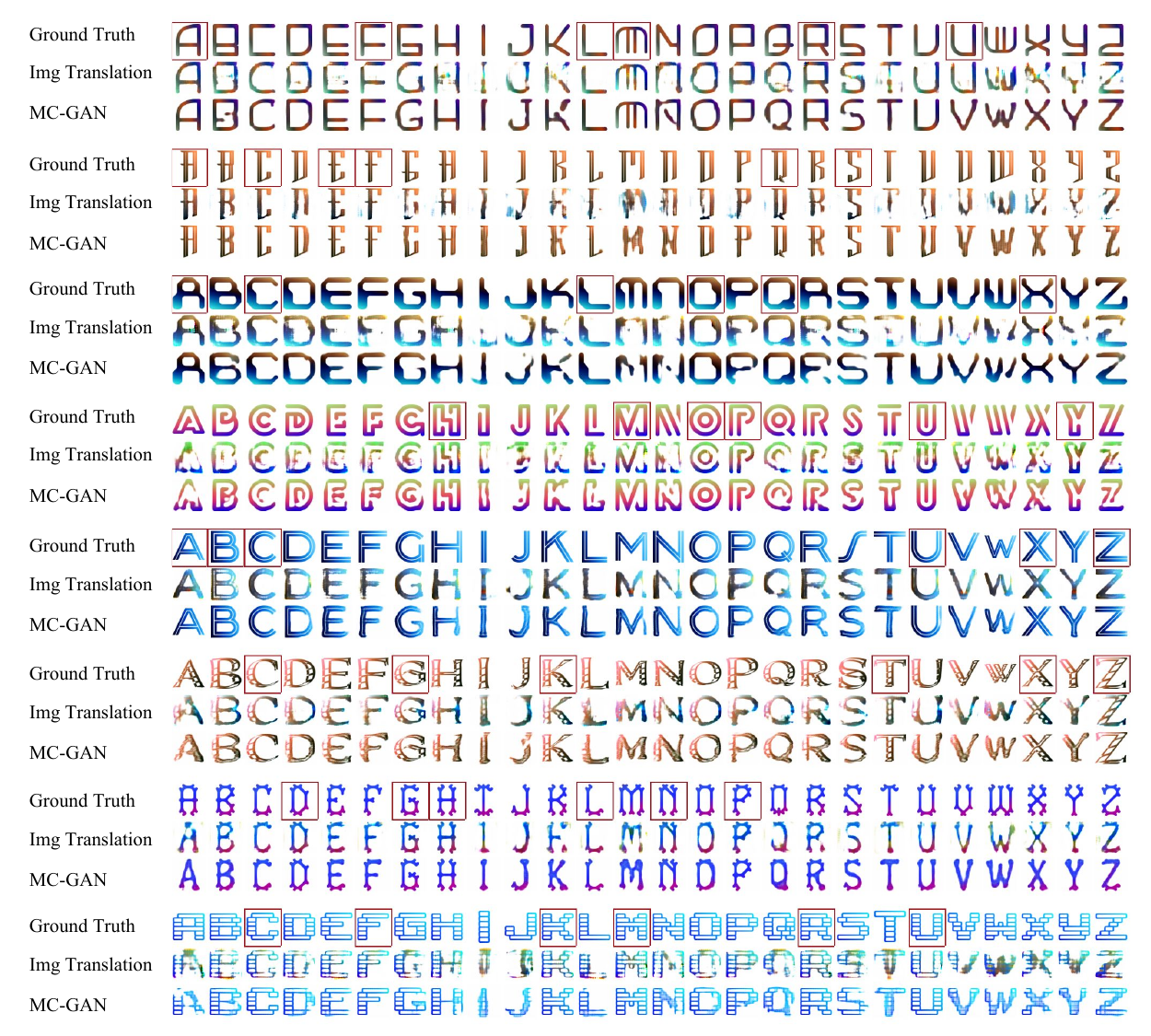}
\caption{Comparison between image translation and our end-to-end multi-content GAN on our synthetic color font data set. For each example, ground truth and given letters are shown in the \textbf{1st row}, image translation outputs in the \textbf{2nd row} and MC-GAN in the \textbf{last row}.} 
\label{fig:color-grad}
\vspace{30em}
\end{figure*}

In this section, we compare our end-to-end multi-content GAN approach with the image translation method discussed in Section~\ref{sec:image-trans}. In Figure~\ref{fig:color-grad}, we demonstrate the results on multiple examples from our color font data set where we have applied random color gradients on the gray-scale glyph outlines. By looking at the nearest neighbor examples, we have made sure that the fonts shown in this experiment were not used during training of our Glyph Network.

Given a subset of color letters in the input stack of GlyphNet with dimension $1\times 78 \times 64\times 64$ including RGB channels, we generate all 26 RGB letters from the pre-trained GlyphNet on our color font data set. Results are denoted as ``Image Translation'' in Figure~\ref{fig:color-grad}. Our MC-GAN results are outputs of our end-to-end model fine-tuned on each exemplar font. The image translation method cannot generalize well in transferring these gradient colors at test time by observing only a few examples although other similar random patterns have been seen during training.

\section{Conclusion and Future Work}
We propose the first end-to-end approach to synthesizing ornamented glyphs from a few examples. Our method takes a few example images as an input stack and predicts coarse shape and fine ornamentations for the remaining glyphs. We train two networks: one for the shape and one for the texture, and demonstrate that by training them jointly, we can produce results that are strongly preferred by users over existing texture transfer approaches that focus on glyphs. 
A surprising discovery of this work is that one can efficiently leverage GANs to address a multi-content style transfer problem. In many practical settings, however, fonts need to be generated at extremely high resolution, motivating extensions to this approach such as hierarchical generation or directly synthesizing smooth vector graphics. In the future, we would also like to explore other problems where content has to be stylized consistently from a few examples. For example, modifying a particular human face (style) to have a specific expression (content), consistent stylization of shapes such as emoticons, or transferring materials to consistent sets of objects such as clothing or furniture.

\vspace{0.3cm}

\textbf{Acknowledgement} Authors would like to thank Elena Sizikova for collecting fonts used in the $10\mathrm{K}$ font data set.

{\small
\bibliographystyle{ieee}
\bibliography{egbib}
}

\appendix
\section{Appendix}
\subsection{Network Architectures}
\label{sec:net-arch}
We employ our generator (encoder-decoder) architecture based on the image transformation network introduced in~\cite{johnson2016perceptual} and discussed in~\cite{isola2016image}. We represent a Convolution-BatchNorm-ReLU consisting of k channels with \texttt{CRk}, a Convolution-BatchNorm layer with \texttt{Ck}, a Convolution-BatchNorm-ReLU-Dropout with \texttt{CRDk}, and a Convolution-LeakyReLU with \texttt{CLk}. In the above notations, all input channels are convolved to all output channels in each layer. We also use another Convolution-BatchNorm-ReLU block in which each input channel is convolved with its own set of filters and denote it by \texttt{CR\textsuperscript{26}k}, where $26$ shows the number of such groups. Dropout rate during training is $50\%$ while ignored at test time. Negative slope of the Leaky ReLU is also set to 0.2.  

\subsubsection{Generators Architecture}
Our encoder architecture in GlyphNet is:
\texttt{CR\textsuperscript{26}26-CR64-CR192-CR576-(CRD576-C576)}-\texttt{(CRD576-CR576)-(CRD576-C576)} where convolutions are down-sampling by a factor of $1-1-2-2-1-1-1-1-1-1$, respectively, and each \texttt{(CRD576-C576)} pair is one ResNet Block.

The encoder in OrnaNet follows a similar network architecture except for in its first layer where the \texttt{CR\textsuperscript{26}26} has been eliminated.

The decoder architecture in both GlyphNet and OrnaNet is as follows:
\texttt{(CRD576-C576)-(CRD576-C576)-(CRD576-C576)} - \texttt{CR192-CR64} each up-sampling by a factor of $1-1-1-1-1-1-2-2$, respectively. Another Convolution layer with 26 channels followed by a \texttt{Tanh} unit is then applied in the last layer of the decoder.

\subsubsection{Discriminators Architecture}
As partially illustrated in Figure~\ref{fig:glyph}, our GlyphNet and OrnaNet discriminators, $D_1$ and $D_2$, consist of a local and global discriminator where weights of the local discriminator is shared with the latter. The local discriminator consists of \texttt{CL64-CL128} followed by a convolution mapping its $128$ input channels to one output. Convolutions here are down-sampling by a factor of ${2-1-1}$, respectively. The global discriminator has two additional layers before joining the layers in the local discriminator as \texttt{CR52-CR52} each down-sampling by a factor of 2. Receptive field size of our local discriminator is 21 while global discriminator covers a larger area than the 64 pixels in the image domain, and thus can capture a global information from each image.

\end{document}